\documentclass{article}

\usepackage{PRIMEarxiv}

\usepackage[utf8]{inputenc} % allow utf-8 input
\usepackage[T1]{fontenc}    % use 8-bit T1 fonts
\usepackage{hyperref}       % hyperlinks
\usepackage{url}            % simple URL typesetting
\usepackage{booktabs}       % professional-quality tables
\usepackage{amsfonts}       % blackboard math symbols
\usepackage{nicefrac}       % compact symbols for 1/2, etc.
\usepackage{microtype}      % microtypography
\usepackage{lipsum}
\usepackage{fancyhdr}       % header
\usepackage{graphicx}       % graphics
\graphicspath{{media/}}     % organize your images and other figures under media/ folder
\usepackage{tabularx} % Para tablas ajustables

\title{Fedetated learning in low-resource settings: A chest imaging study in Africa - Challenges and lessons learned}
\author{Jorge Fabila \\
Dept. de Matemàtiques i Informàtica \\
Universitat de Barcelona\\
Barcelona\\ Spain\\
jorge\_fabila@ub.edu
   \And
Lidia Garrucho\\
Dept. de Matemàtiques i Informàtica \\
Universitat de Barcelona\\
Barcelona\\\\ Spain
   \And
Víctor M. Campello\\
Dept. de Matemàtiques i Informàtica \\
Universitat de Barcelona\\
Barcelona\\\\ Spain
   \And
Carlos Martín-Isla\\
Dept. de Matemàtiques i Informàtica \\
Universitat de Barcelona\\
Barcelona\\\\ Spain
   \And
Karim Lekadir\\
Dept. de Matemàtiques i Informàtica \\
Universitat de Barcelona\\
Institució Catalana de Recerca i Estudis Avançats (ICREA)\\ Barcelona\\ Spain
}

\begin{document}
\maketitle

\begin{abstract}
This study investigates the application of Federated Learning (FL) in low-resource settings, focusing on its potential for tuberculosis (TB) diagnosis through chest X-ray imaging in Africa. Traditional centralized AI models face significant challenges related to data privacy regulations and limited access to diverse datasets, making FL an appealing alternative. By allowing institutions to collaborate on model training without sharing raw patient data, FL offers a solution that preserves privacy while leveraging distributed computing resources.  

The research involved collaborations with hospitals and research centers across several African countries, including Ethiopia, Ghana, The Gambia, Mozambique, Nigeria, the Democratic Republic of the Congo, Senegal, and Uganda. While most sites used locally collected datasets, Ghana and The Gambia relied on publicly available data due to limitations in obtaining local samples. The goal was to assess the feasibility of FL in real-world healthcare applications by comparing models trained locally at each hospital with a federated model trained collectively across institutions. 

Despite its advantages, the implementation of FL in low-resource environments like sub-Saharan Africa remains challenging. Several technological and structural barriers were identified. Many hospitals lacked the necessary infrastructure to support FL, with unreliable network connectivity further complicating deployment. Additionally, concerns over data privacy persist, as even though FL does not share raw data, risks such as model inversion attacks could still potentially expose sensitive information. Another major obstacle was the lack of digital health literacy among medical staff and policymakers, which hindered adoption. Furthermore, fragmented healthcare systems and weak AI regulations across Africa made it difficult to establish clear policies for data governance and cross-border collaborations. There was also resistance to AI-driven healthcare solutions, as institutions were reluctant to share model updates, fearing a loss of control over their data.

Ultimately, this research provides real-world evidence that Federated Learning offers a promising solution for advancing medical AI in low-resource settings. By addressing data scarcity, privacy concerns, and infrastructure challenges, FL has the potential to democratize AI-driven healthcare innovations, enabling hospitals in underserved regions to benefit from cutting-edge technology. However, its widespread adoption will require continued research and development, with a focus on improving scalability, usability, and regulatory frameworks to ensure long-term success.

\end{abstract}

% keywords can be removed

\section{Introduction}
With the increasing demand for privacy-preserving medical AI, traditional centralized approaches face significant challenges in handling sensitive data. Federated learning (FL) offers a decentralized solution by enabling models to be trained across multiple devices or servers without sharing raw data [1] [2]. %\cite{McMahan2016} \cite{tian_li}. 
Instead of transferring raw data to a central server, only model updates --such as weights and parameters-- are exchanged and securely aggregated [3] [4].%\cite{sohan2023} \cite{moshawrab2023}. %This approach ensures data privacy and leverages distributed computational resources . 
By minimizing the risk of exposing sensitive information, FL aligns with growing regulatory requirements, such as GDPR, which emphasize the importance of secure and private data handling [5] [6]. %\cite{dayan2021} \cite{sabina}.
As a result, FL offers a transformative solution to key challenges related to data privacy, security, and scalability [7]. %\cite{yang2019}. 
FL has been used successfully in different fields of applied artificial intelligence in medicine [8] [9] [10]. % \cite{linardos2022} \cite{rehman2023} \cite{soltan2024}.

In addition, FL fosters collaboration between organizations, allowing the grouping of insights from various datasets to create more robust and generalizable models. This is particularly valuable in contexts where data is scarce or incomplete (for example, only a few categories are available), or where limited computational resources hinder the adoption of advanced AI technologies [11].%\cite{Wen2023}.  
Although some models, such as transformers, are resource-intensive and may pose challenges for devices with limited capabilities, FL frameworks offer strategies to address these limitations [12]. %\cite{ZHANG2021106775} . 
For instance, techniques like model compression (quantization, pruning, or distillation), hierarchical or split learning (offloading resource-heavy operations to more capable nodes or a central server), and dynamic workload distribution (assigning simpler tasks to less powerful devices) enable FL to function effectively in resource-constrained environments [13] [14]. %\cite{LI2020106854} \cite{sannara}.  %These adaptations allow FL to promote inclusion and collaboration, even in scenarios with heterogeneous hardware resources \cite{LI2020106854} \cite{sannara}.
A compelling example is that in certain regions in Africa, that face significant challenges in healthcare delivery due to inadequate infrastructure and restricted access to cutting-edge medical technologies. These challenges are not uniform across the continent, as countries and regions exhibit substantial diversity in their healthcare systems and resources. For instance, rural areas in sub-Saharan Africa often encounter barriers related to lack of infrastructure and access to healthcare services [15] %\cite{Ade-Ibijola2023}.% FL could provide a means to bridge these gaps, allowing healthcare institutions to collaborate while maintaining data privacy.
%==========================

Despite its numerous advantages, Federated Learning (FL) has yet to achieve widespread adoption, largely due to several persistent challenges that prevents its successful implementation [16]. %\cite{MAL-083}.
One of the most pressing issues is system heterogeneity, as FL must operate across diverse computational environments with different hardware capabilities, network conditions, and software infrastructures. This heterogeneity makes it difficult to ensure consistent model performance across different institutions, particularly in regions with limited technological resources.

%Another critical challenge is data privacy and security \cite{soltan2024}. 
While FL inherently reduces the need to share raw data by transmitting only model updates, this does not entirely eliminate privacy risks. Techniques such as adversarial attacks, model inversion, and inference attacks could still be used to extract sensitive information from shared updates, significantly threatening to patient confidentiality [17]. %\cite{Liu2022}. 
Moreover, the growing regulatory landscape surrounding the use of AI in healthcare further complicates matters, as many jurisdictions impose strict guidelines on the sharing of models trained on sensitive medical data, even when direct data exchange does not occur. % \cite{deyan2021}.

Connectivity issues represent another major barrier to FL deployment, particularly in low-resource settings. Unstable internet connections, network restrictions imposed by local VPNs, and the technical complexity of configuring FL frameworks rise challenges for institutions that lack dedicated IT support. %\cite{tian_li}. 
Establishing and maintaining a secure and reliable communication infrastructure between federated nodes is crucial for the success of FL, yet it remains one of the most difficult aspects to manage, especially in regions where digital infrastructure is underdeveloped.

FL has demonstrated promising results in experimental settings [18] [19], %\cite{Yao2021} \cite{He2022}, 
its scalability, reliability, and practical impact in real-world applications remain uncertain. Most research has been conducted in simulated environments [20] [21], % \cite{He2022ClassWise} \cite{Dong2022},
leaving open questions about how FL models perform when exposed to highly heterogeneous datasets, real-world operational constraints, and imperfect data quality [24]. %\cite{soltan2024scalable}. 
Without further validation in practical healthcare scenarios, stakeholders may remain hesitant to adopt FL at scale [23]. % \cite{yan}.

To address these gaps, this study investigates the application of FL in a realistic, resource-constrained setting, focusing on chest radiograph images for tuberculosis diagnosis across multiple hospitals in Africa. Through this research, we encountered a series of practical challenges, including data scarcity, model performance disparities across regions, and infrastructure limitations. By systematically analyzing these obstacles, we propose potential solutions aimed at improving the viability and efficiency of FL in environments where traditional AI methods struggle due to privacy concerns and fragmented data availability. Our findings lay the groundwork for enhancing the adoption of FL in medical AI, particularly in regions with limited access to centralized computing resources, offering a pathway toward more inclusive and privacy-preserving healthcare innovations.

\section*{Methods}

\subsection*{Datasets}

A total of eight research centers from eigth Sub-Saharan African (SSA) countries, including Ethiopia, Ghana, The Gambia, Mozambique, Nigeria, Democratic Republic of the Congo, Senegal and Uganda, have collaborated in this study. 
All of them used local datasets for training, except for The Gambia and Ghana, where no retrospective datasets where available for training. Instead of discarding both research centers from the study, we used a publicly available data set [24] %\cite{Rahman2020}
to be able to test the rest of the federated pipeline. 
The public dataset used in both centers was gathered by researchers from the Qatar University, the University of Dhaka, a group of scientists from Malaysia and medical doctors from Hamad Medical Corporation, comprising a total of 4,200 chest X-ray (CXR) images in PNG format and categorized into two groups: tuberculosis and normal.

\begin{figure}
    \centering
    \includegraphics[width=1\linewidth]{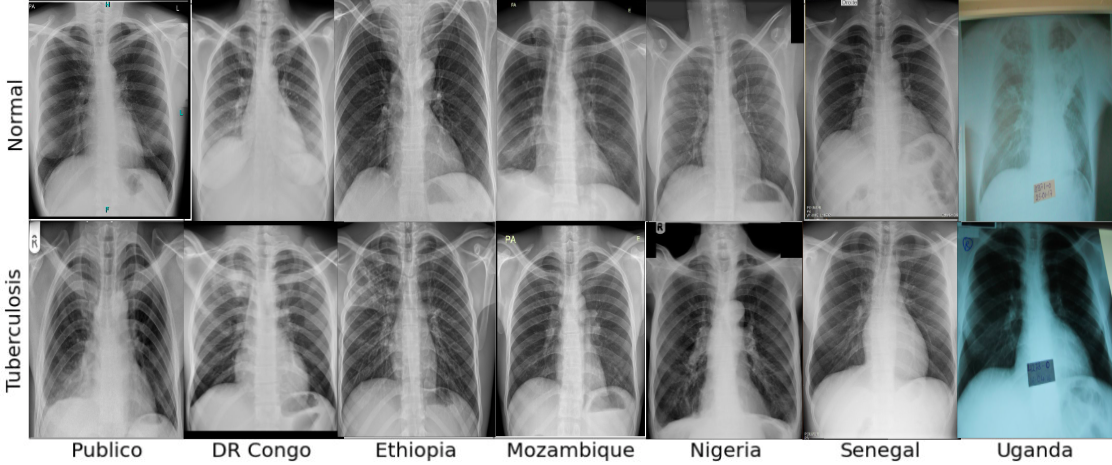}
    \caption{Top: Samples of normal X-rays images.  Bottom: Samples of tuberculosis images.
    From left to right: Public dataset, DR of the Congo, Ethiopia, Mozambique, Nigeria, Senegal, Uganda}
    \label{fig:muestras}
\end{figure}

Illustrative examples of the X-ray scans available in each center for healthy individuals and for patients diagnosed with tuberculosis  are shown in Figure \ref{fig:muestras}.

%To maintain the integrity of the FL set-up, this public dataset was partitioned and distributed across the participating nodes to simulate local datasets. 
This approach ensured that all nodes contributed to the training process with their own local data set. A detailed summary of all partners and the provided data sets is shown in Table \ref{tab:datasets_1}. 
Table \ref{tab:datasets_2} provides a breakdown of the content of the data set and the characteristics of the population studied in each center. The datasets were divided into training and testing sets based on a random split, 70\% for training, 20\% for validation and 10\% for testing.
Figure 2 shows the distributions of labels per dataset. In particular, one can observe the similarities between the cohorts distributions hosted in the Republic of The Gambia and in Ghana, which used splits of the same public dataset. All of them are highly imbalanced. For example, the Gambian dataset contains 83\% negative cases, while the Congolese dataset has 78\% positive cases.

Looking at Figure \ref{fig:muestras}, a crucial factor to consider in this study is the quality of medical images obtained from Uganda, which was noticeably lower compared to datasets from other regions. Several factors likely contributed to this discrepancy, including differences in imaging equipment, variations in acquisition protocols, and potential inconsistencies in image preprocessing workflows. Such limitations can introduce noise, artifacts, and reduced feature clarity, making it more challenging for the model to extract relevant patterns and generalize effectively. However, we acknowledge that low-quality images can pose significant challenges for model training and rather than viewing this limitation solely as a drawback, we recognize it as an opportunity to rigorously test the resilience and adaptability of our federated learning methodology. If a federated model is able to achieve robust and generalizable performance despite variations in image quality, this would provide strong evidence that FL can be an effective solution in real-world, resource-limited healthcare settings. Evaluating the impact of suboptimal data quality within the federated framework allows us to assess the model’s ability to integrate heterogeneous datasets and determine whether high-performing AI models can still be obtained even when some participating institutions contribute lower-resolution or noisier images.

%\begin{table}[h!]
%\centering
%\caption{Distribution of cohorts used in this study across participating centers.}
%\begin{tabular}{llcc}
%\toprule
%\textbf{Country} &\textbf{Cohort} & \textbf{Dataset} & \textbf{Sample size} \\
%\midrule
%DR of the Congo & Université de Kinshasa & Local & 211 \\
%Ethiopia & Jimma University & Local & 133 \\
%Ghana & Delft Imaging Ghana LTD & Public & 250 \\
%Mozambique  & Fundaçao Manhiça & Local & 205  \\
%Nigeria  & Nigerian institute of medical research & Local & 1726 \\
%Rep. of The Gambia & Medical Research Council Unit The Gambia & Public & 250 \\
%Senegal  & Université Iba Der Thiam de Thies   & Local & 98 \\
%Uganda & Infectious diseases research collaboration LTD  & Local & 507 \\
%\bottomrule
%\end{tabular}
%\label{tab:datasets_1}
%\end{table}

\begin{table}[ht]
\centering
\begin{tabular}{lllll}
\toprule
\textbf{Country} & \textbf{Acronym} & \textbf{Dataset} & \textbf{Origin}  & \textbf{Sample size} \\
\midrule
DR of the Congo & DRG & Local & Université de Kinshasa &  211 \\
Ethiopia & ETH & Local & Jimma University Medical Center  & 133 \\
Ghana & GHA & Public & Public Dataset [9]  & 250 \\
Mozambique & MZQ & Local & Fundação Manhiça &  205  \\
Nigeria  & NIG & Local & Nigerian Institute of Medical Research & 1726 \\
The Gambia & GAM &  Public & Public Dataset [9] & 250 \\
Senegal  & SEN & Local & Université Iba Der Thiam de Thies   & 98 \\
Uganda & UGN & Local & China-Uganda Friendship Hospital \& Mulago Hospital  & 507 \\
\bottomrule
\end{tabular}

\caption{Summary of datasets by country and origin.}
\label{tab:datasets_1}
\end{table}

%Congo, Etiopia, Gambia, Ghana, Mozambique, Nigeria, Senegal Uganda

\begin{table}[h!]
\centering
%\scriptsize
\begin{tabular}{lllccccc|c|c|}
\toprule
\textbf{Country}&\textbf{Scanner model} &  \textbf{Patients} & \textbf{Sex} & \textbf{Age} & \textbf{Ethnicity}\\
& & & \textbf{M - F (\%)} & \textbf{Mean Std Min Max} & \\
\midrule
DR Congo  & & 211 & 58.45 - 41.55 & 39 ; 18 ; 1 ; 98 & No further info\\
Ethiopia  &  DigiEye 330  (Mindray) & 133 &  No further info &  No further info & No further info\\
Ghana & Kodak Point-of-Care 260 ; unknown & 250  & No further info &  No further info &  No further info\\
Mozambique   & & 205 & & & \\
Nigeria  &   & 1726 &  & &\\
The Gambia   &  Kodak Point-of-Care 260 ; unknown & 250 &  No further info &  No further info &  No further info \\
Senegal  & Camstream & 98 & 40 - 60 &30-59 & Black \\
%Uganda &  BLD-150RK (Listem) & 507  & 61 - 39 &Mean: 35.0; Standard deviation: 9.61; Min: 18 Max: 64 & Black\\
Uganda &  BLD-150RK (Listem) & 507  & 61 - 39 & 35.0; 9.61; 18 ; 64 & Black\\
\bottomrule
\end{tabular}

\caption{Summary of datasets contents (population characteristics).}
\label{tab:datasets_2}

\end{table}

%Canon  Cus-X100g  Canon Medical Systems
%Philips Affinity AFFINITY 709 /795210 PHILIPS, USA

\begin{figure}
\centering
\includegraphics[width=13cm]{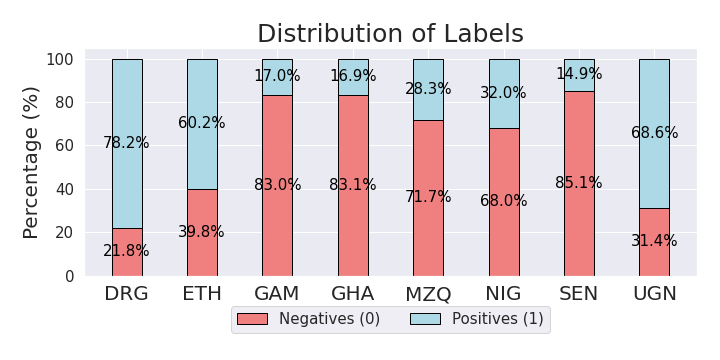}
\caption{Distribution of labels for all of the datasets.}
\label{fig:samples}
\end{figure}
The dataset used in this study is private and cannot be publicly shared due to data protection regulations and confidentiality agreements with the participating institutions. Access to the data is restricted to ensure compliance with ethical guidelines and privacy policies. Researchers interested in potential collaborations may contact the corresponding author to explore possibilities within the constraints of data-sharing agreements.

\subsection*{Training}

We developed a FL platform using the Flower library (flwr) [25],%\cite{beutel2020},
PyTorch [26], %\cite{ansel2024},
and PyTorch Lightning [27].  %\cite{falcon2019}. 
Flower serves as the communication framework between the central server and the various nodes via a virtual private network, ensuring a secure connection. Specifically, Flower coordinates the FL process by handling the exchange of model weights between the central server and the nodes. Each node trains a local copy of the model, which is trained using its local data. After training, the node sends the updated model weights back to the central server, where they are aggregated to improve the global model. This process is repeated iteratively until convergence is achieved. Flower handles not only the secure transfer of model weights but also manages tasks such as task scheduling, resource management, and model updates, ensuring the synchronization of training across nodes in a decentralized manner.
The optimizer used was Adam with a learning rate scheduler (ReduceLROnPlateau) that dynamically adjusts the learning rate based on the training performance, avoiding the stagnation of the training by reducing by a factor of 0.1 every 10 validation steps if no improvement in the validation loss is observed. Our framework includes different models and variants of them, but for this study, we use ResNet-50 [28],% \cite{he2016}, 
as it provides a good trade-off between accuracy and resource usage.

We trained a federated learning (FL) model by first training individual models on each site's local dataset. Each local model, trained for 100 epochs, represents our baseline.

Afterward, federated training was performed by individually training each node for 10 epochs, with the model weights then sent to the central server. This process, referred to as a "round," was repeated until all nodes had completed their individual 10-epoch training cycles. The server aggregated the weights from all local models to create a global model after each round. %, and this global model was then fine-tuned for additional epochs.
The training process was conducted over a total of 20 rounds, with each round consisting of 10 individual epochs.

To assess the generalizability of the FL model, we tested each trained model on the testing datasets of all other centers, ensuring the evaluation was performed across different datasets to validate the model's robustness. As a comparison, we also trained single-center models independently and compared their performance to that of the federated model. This approach allowed us to evaluate the benefits of federated learning in terms of model generalizability and performance across diverse datasets.

For performance assessment, we used sensitivity, specificity, balanced accuracy, and the area under the receiver operating characteristic curve (ROC AUC). To briefly recall the key concepts, the metrics are defined as follows. Sensitivity, or True Positive Rate, measures the model's ability to correctly identify positive instances. Specificity, or True Negative Rate, assesses how well the model classifies negative cases. Balanced Accuracy is calculated as the average of sensitivity and specificity, offering a more reliable metric, especially in the context of imbalanced datasets. Finally, the Area Under the Receiver Operating Characteristic Curve (ROC AUC) provides a comprehensive evaluation of the model's discriminative power across various classification thresholds. %, which are defined as follows.

\section{Results}
To begin, we utilized a public dataset to evaluate the classifier's performance and determine the minimum amount of data required to obtain a robust model. This was a critical step because the data collected from the hospitals participating in this study were limited. Understanding the minimum data requirements beforehand allowed us to estimate how much data we needed to request from each hospital to achieve satisfactory model performance.

\begin{figure}
    \centering
    \includegraphics[width=0.49\linewidth]{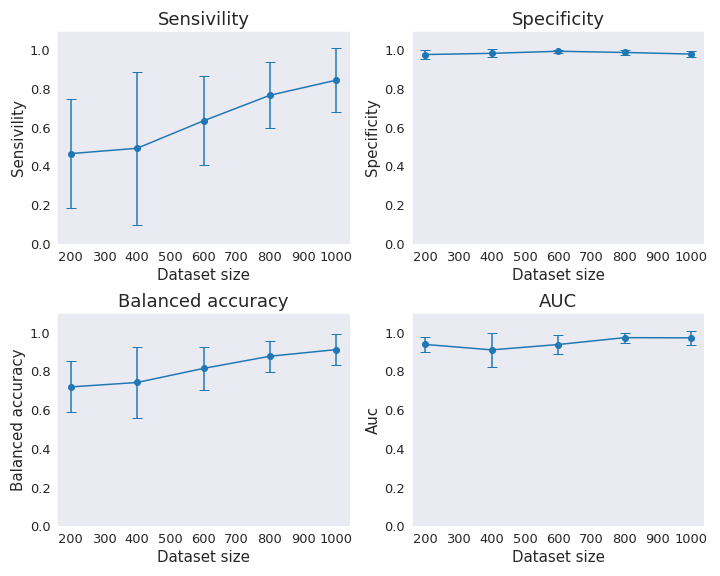}\includegraphics[width=0.49\linewidth]{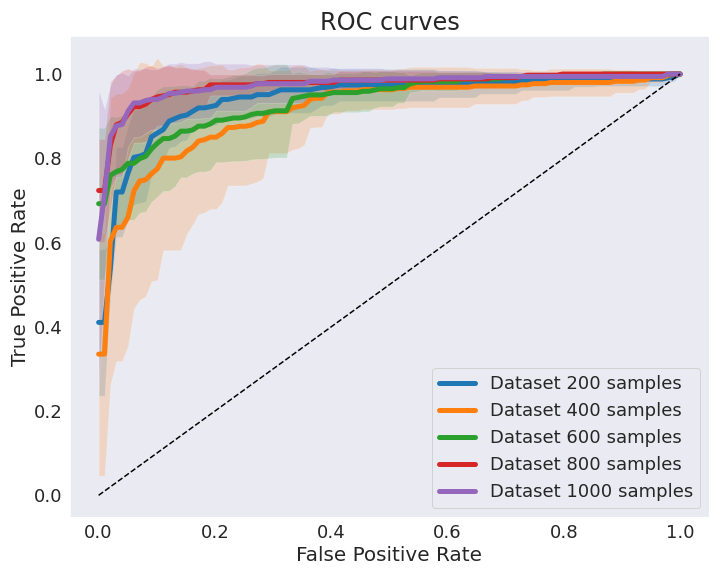}

    \caption{Left: Sensitivity, specificity and balanced accuracy as dataset size increases (Public dataset). Right: ROC-AUC as dataset size increases (Public dataset)}
    \label{fig:metrics}
\end{figure}

Using the public dataset, we trained a ResNet-50 classifier on subsets of varying sizes, ranging from 200 to 1,000 samples, increasing in increments of 200 images. For each subset, 80\% of the data was used for training, 10\% for validation, and the remaining 10\% for testing. Figure \ref{fig:metrics} illustrates how sensitivity, specificity, and accuracy improved as the dataset size increased. The results indicate that datasets with more than 600 images were sufficient to produce a model with good performance.

This analysis also highlights a significant limitation: individual hospitals alone could not generate robust models due to their limited data availability (see Table \ref{tab:datasets_1}). Consequently, federated learning (FL) offers a practical solution to this issue by enabling collaborative model training across multiple hospitals. This approach allows for the creation of robust models that would otherwise be unfeasible using the data from a single institution.

%To initiate our study, we first employed a publicly available dataset to assess the classifier’s performance and determine the minimum data requirements necessary to develop a robust and reliable model. This preliminary step was crucial, given the limited availability of medical data from the hospitals participating in this research. By establishing a baseline for minimum dataset size, we were able to estimate the data contribution needed from each hospital to achieve an acceptable level of model performance, ensuring that the federated learning framework could be implemented effectively.  

%Using this public dataset, we trained a ResNet-50 classifier on progressively larger subsets, ranging from 200 to 1,000 samples, increasing in increments of 200 images. The dataset was divided into 80\% for training, 10\% for validation, and 10\% for testing. As illustrated in Figure \ref{fig:metrics}, key performance metrics—including sensitivity, specificity, and accuracy—showed consistent improvement as the dataset size increased. The results indicated that models trained on datasets containing at least 600 images exhibited stable and satisfactory performance, suggesting this as a threshold for reliable classification.  

%However, this analysis also exposed a critical limitation: individual hospitals, 
Due to the restricted data availability in individual hospitals, were incapable of training high-performing models independently (see Table \ref{tab:datasets_1}). The inability to aggregate a sufficient volume of training data within a single institution significantly limits the model robustness and generalization. This underscores the necessity of FL as a viable solution, as it enables collaborative model training across multiple institutions without requiring data to be centralized. % which would otherwise be unattainable using data from a single institution alone.  

%\begin{figure}
%    \centering 
%\includegraphics[width=0.5\linewidth]{comparison_metrics_1.png}
%    \caption{Sensitivity, specificity and balanced accuracy as dataset size increases (Public dataset).}
%    \label{fig:metrics}
%\end{figure}

%\begin{figure}
%    \centering
%\includegraphics[width=0.5\linewidth]{ROC_Curve_Comparison.png}
%    \caption{ROC-AUC as dataset size increases (Public dataset).}
%    \label{fig:roc}
%\end{figure}

Figure \ref{fig:metrics} provides a comprehensive summary of the models' performance based on the respective evaluation metrics presented in each figure. When analyzing sensitivity and specificity, a striking trend emerges: nearly all models—except for those trained on the Nigeria dataset and the federated model—exhibit extreme values, with sensitivity and specificity scores predominantly at either 1 or 0, leaving no room for intermediate results. This behavior reflects the underlying class imbalance present in the datasets.

For example, datasets from the Democratic Republic of the Congo, Ethiopia, and Uganda contain a significantly higher proportion of TB-positive cases compared to negative ones. As a consequence, models trained on these datasets tend to overfit to the positive class, classifying almost all samples as TB-positive. This phenomenon is particularly evident in models trained on the Ethiopia and Uganda datasets, where specificity drops to 0, meaning they fail to correctly identify any negative cases. Conversely, models trained on datasets from other regions—excluding Nigeria and the federated model—exhibit a specificity of 1, indicating that they classify all cases as negative, completely ignoring the presence of positive samples.

A similar pattern is observed when examining sensitivity. Models trained on the Ethiopia and Uganda datasets reach a sensitivity score of 1, meaning they correctly classify all positive cases but fail to recognize negative ones. In contrast, the remaining models, with the exception of those trained on Nigeria’s dataset and the federated model, display sensitivity values close to 0, further confirming the severe overfitting to either class.

This analysis clearly demonstrates the limitations of training models on individual datasets. With the exception of the Nigeria model, all locally trained models exhibit poor generalization, failing to adapt when tested on external datasets. The Nigeria model stands out as an exception, achieving 90\% sensitivity and 85\% specificity when evaluated on its own test set. However, its performance deteriorates significantly when applied to different datasets. For instance, when tested on Mozambique’s dataset, its specificity drops to 26\% and sensitivity to 67\%, highlighting its lack of robustness in cross-regional generalization. This discrepancy is further reflected in its accuracy, which reaches 87.52\% on its own dataset but declines sharply to 44\% when tested on Mozambique’s data.

\begin{figure}
    \centering
    \includegraphics[width=0.32\linewidth]{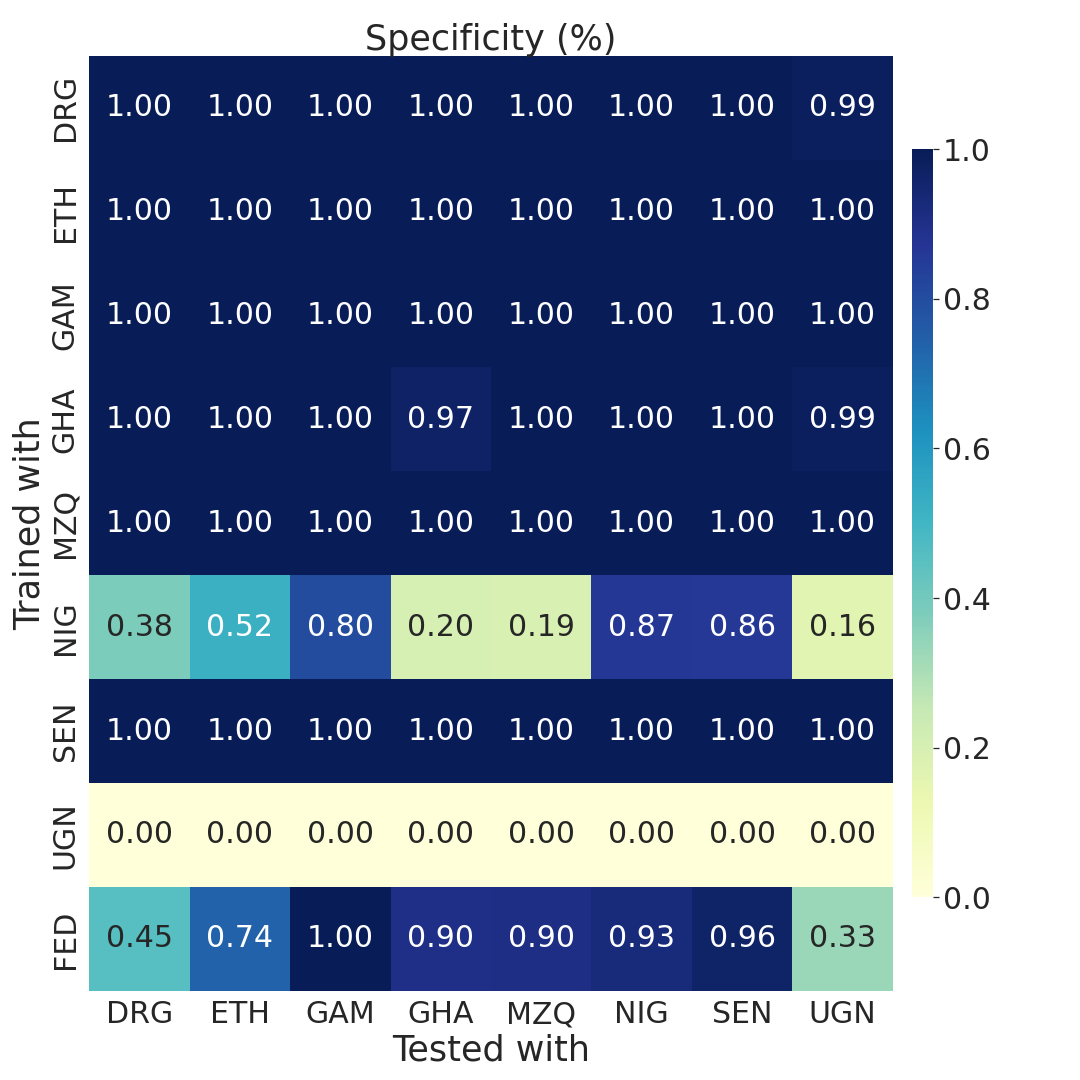}
\includegraphics[width=0.32\linewidth]{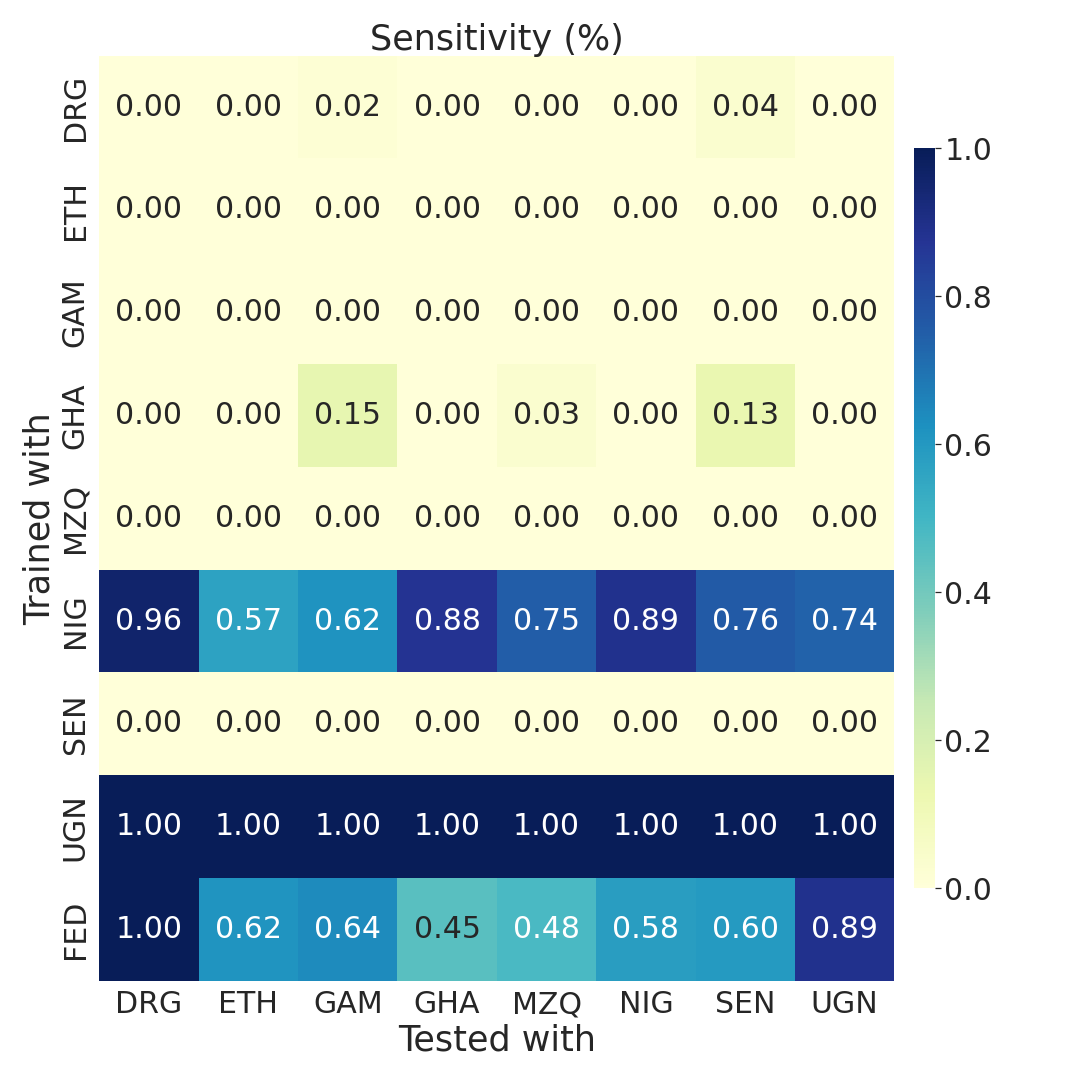}
\includegraphics[width=0.32\linewidth]{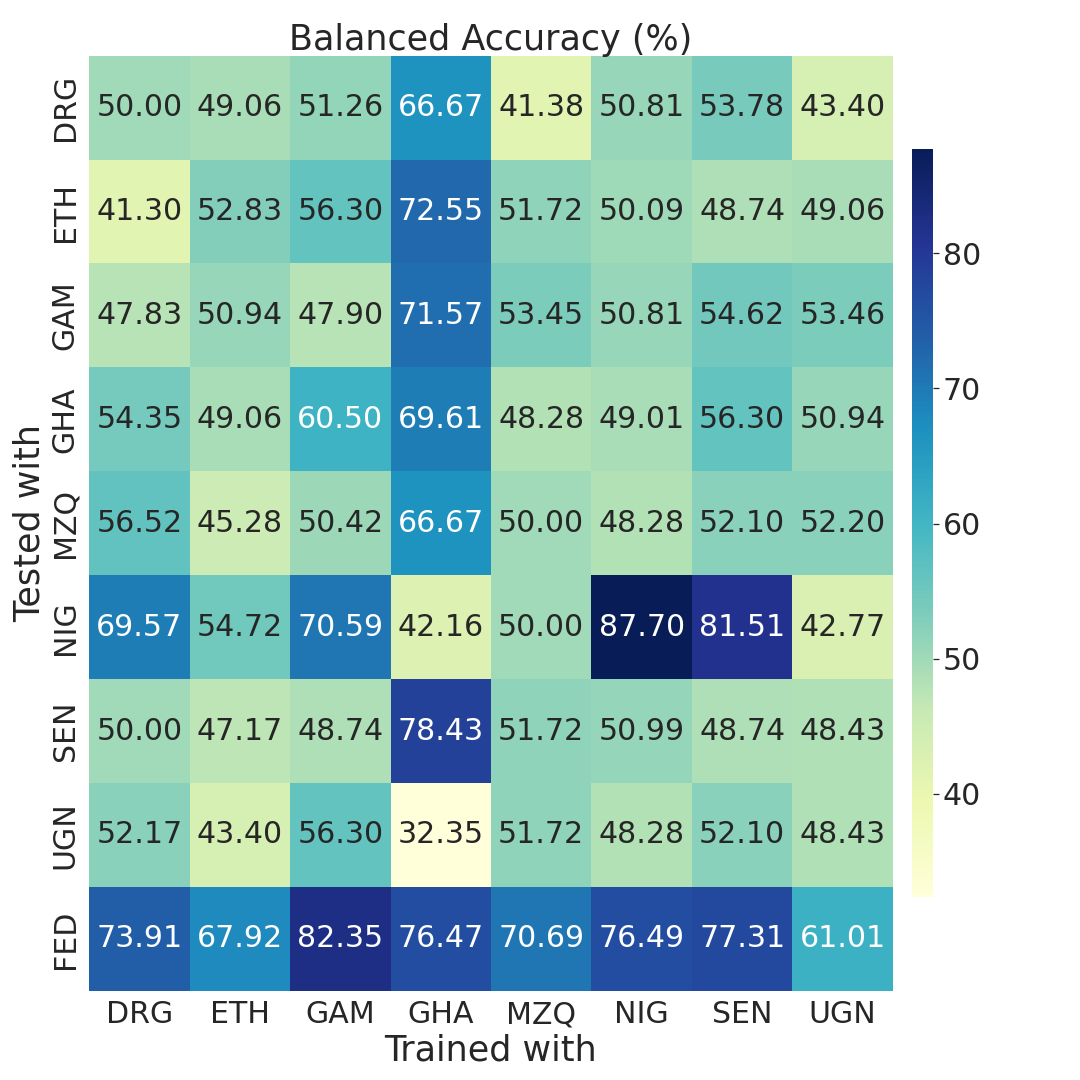}

    \caption{Specificity, sensitivity and balanced accuracy of baseline regional models compared with federated model.}
    \label{fig:specificity}
\end{figure}

In contrast, the federated model exhibits superior performance across all three evaluation metrics, demonstrating a significantly improved ability to generalize across different datasets. This enhancement is particularly noticeable in balanced accuracy, as illustrated in the last row of Figure \ref{fig:specificity}. The federated approach effectively mitigates the issues caused by class imbalances and regional overfitting, leading to the development of a more robust and generalizable model that performs consistently across diverse datasets.

%\begin{figure}
%    \centering
%    \includegraphics[width=0.5\linewidth]{especificidad_RX.png}
%    \caption{Specificity of baseline regional models compared with federated model.}
%    \label{fig:specificity}
%\end{figure}

%\begin{figure}
%    \centering
%    \includegraphics[width=0.5\linewidth]{sensitivity_RX.png}
%    \caption{Sensitivity of baseline regional models compared with federated model.}
%    \label{fig:sensitivity}
%\end{figure}
%\begin{figure}
%    \centering
%    \includegraphics[width=0.5\linewidth]{ac_RX.png}
%    \caption{Balanced accuracies of baseline regional models compared with federated model.}
%    \label{fig:accuracies}
%\end{figure}

%\begin{figure}
%    \centering
%    \includegraphics[width=0.35\linewidth]{ROC_Curve_200.png}

%    \includegraphics[width=0.35\linewidth]{ROC_Curve_400.png}
%    \includegraphics[width=0.35\linewidth]{ROC_Curve_600.png}
%    \includegraphics[width=0.35\linewidth]{ROC_Curve_800.png}
%    \includegraphics[width=0.35\linewidth]{ROC_Curve_1000.png}
    
%    \caption{Caption}
%    \label{fig:enter-label}
%\end{figure}

%\begin{table}[h!]
%\centering
%\begin{tabular}{|c|c|c|c|c|}
%\hline
%\textbf{Dataset} & \textbf{Type} & \textbf{Test with local dataset}  & \textbf{Test with global dataset}   \\
%\hline
%The Gambia &  Public & 68.53  & 91.13 \\
%\hline
%Senegal  & Local & 66.29 & 98.87 \\
%\hline
%Ghana & Public & 65.29 & 71.91 \\
%\hline
%Nigeria & Local & 65.16 & 71.91 \\
%\hline
%Gabon &  Public & 66.29 & 78.65 \\
%\hline
%Mozambique & Local & 69.66 & 94.38 \\
%\hline
%DR Congo & Local & 38.20 & 87.64 \\
%\hline
%Uganda &  Local & 37.07 & 75.28 \\
%\hline
%Federated & Distributed &  & 75.28 \\
%\hline
%\end{tabular}
%\caption{Number of samples per data set}
%\label{tab:datasets}
%\end{table}
\section*{Discussion}

The results obtained in this study underscore the significance and advantages of Federated Learning (FL) when compared to individually trained models, particularly in settings characterized by heterogeneous and region-specific data distributions. Our findings emphasize that the methodological approach employed in this work is well-suited for real-world applications, demonstrating its feasibility beyond controlled experimental environments.

One of the most critical challenges in medical AI is the tendency of models to overfit when trained on localized datasets, severely limiting their ability to generalize to different geographic regions. This issue was particularly evident in the Nigeria model, which achieved strong performance on its own test set, with 90\% sensitivity and 85\% specificity. However, when tested on external datasets—such as Mozambique’s—the model’s specificity dropped drastically to 26\%, while sensitivity declined to 67\%. This stark contrast highlights the phenomenon of data regionalism, where models become highly attuned to local features, preventing them from adapting effectively to diverse populations. Such limitations are especially problematic in real-world medical applications, where data variability is inevitable, and models must generalize across different demographic and clinical settings.

In contrast, the federated model consistently outperformed individual models across all key evaluation metrics, including sensitivity, specificity, and balanced accuracy. The latter is particularly critical in real-world applications, where imbalanced datasets pose a major challenge. By aggregating insights from multiple institutions while maintaining data privacy, FL effectively mitigates the limitations associated with data scarcity, bias, and regional overfitting. This ability to leverage diverse datasets without compromising patient confidentiality makes FL an attractive solution for healthcare AI applications, particularly in settings where data fragmentation and privacy concerns hinder traditional machine-learning approaches.

The improved generalization capacity of the federated model stems from its ability to integrate representations learned from multiple data domains, allowing it to capture a broader spectrum of clinical variations. This robustness is particularly evident in the results presented in Figure \ref{fig:specificity},  One particularly notable case is the Ugandan dataset, which, despite its low image quality, still contributed to the overall learning process without significantly degrading model performance. This finding underscores the robustness of the federated approach, as it was able to integrate noisier, lower-resolution images while maintaining strong classification accuracy.

This robustness is particularly noteworthy given that deep learning models are typically sensitive to data quality, with lower-resolution inputs often leading to weaker feature extraction and higher classification errors. %However, our results indicate that FL enables a more adaptable and resilient learning process, allowing the global model to leverage information from diverse sources while compensating for variability in image quality. Rather than being hindered by the presence of lower-quality images, the federated model demonstrated stable performance across all participating datasets, reinforcing the idea that heterogeneous data contributions can enhance model generalization rather than compromise it.

A distinctive aspect of this study is its reliance on real-world clinical data collected from multiple institutions, rather than synthetic or laboratory-controlled datasets. While many prior FL studies remain confined to simulated environments, our work provides practical evidence of FL’s effectiveness in real-world applications, reinforcing its viability in environments characterized by regional biases, data scarcity, and inconsistent data quality. The success of our approach highlights not only the technical feasibility of FL but also its potential to democratize access to AI-driven healthcare solutions in resource-limited settings.

By enabling collaborative model training without requiring raw data exchange, FL aligns with strict privacy regulations and safeguards patient confidentiality, making it particularly relevant in contexts where healthcare data is highly fragmented across institutions and geographic regions. This study presents compelling evidence that FL can address some of the most pressing challenges in AI-driven medical research, including data heterogeneity, imbalance, and privacy constraints.

The findings of this study have far-reaching implications for the adoption of AI-based healthcare systems. By demonstrating that FL can overcome key technical and ethical barriers, this research lays the foundation for the development of more equitable, robust, and generalizable AI models in clinical applications. However, despite its promise, technical and infrastructural challenges remain, particularly concerning usability, model optimization, and deployment scalability. These limitations signal the need for further research aimed at refining the user experience, streamlining training processes, and improving the overall implementation of FL-based methodologies in real-world healthcare settings. Table \ref{tab:mitigation} provides a summary of the key challenges encountered and potential solutions proposed.

\begin{table}[h!]
\centering
\renewcommand{\arraystretch}{1.5} % Espaciado entre filas
\setlength{\tabcolsep}{5pt} % Espaciado horizontal en celdas
\begin{tabularx}{\textwidth}{|>{\raggedright\arraybackslash}p{4cm}|X|}
\hline
\textbf{Problem} & \textbf{Possible Solution} \\ 
\hline
Data security and privacy concerns & Provide training on anonymization techniques and federated learning capabilities. Continuously promote benefits of data sharing and new AI tools for society. \\ 
\hline
Technological limitations in SSA & Develop a hybrid platform that accommodates varying scenarios (i.e., federated, cloud-based, centralized). Promote data monetization to increase resources. \\ 
\hline
Lack of digital health literacy & Create extensive training programmes, accessible online resources, and workshops to improve digital health literacy among stakeholders. \\ 
\hline
Fragmented healthcare systems & Engage with healthcare policymakers and healthcare managers from the start to ensure the platform and tools are designed with interoperability in mind. \\ 
\hline
Geopolitical and language barriers & Ensure the platform and training materials are accessible in multiple languages across Secure Service Access (SSA). \\ 
\hline
Scaling-up challenges & Plan for scalability from the beginning, including technical capacity and digital strategy. Use open calls to showcase and optimize scalability. \\ 
\hline
Resistance to data sharing & Implement pilot use cases that show clear benefits for all researchers, such as increased scientific outputs. \\ 
\hline
Resistance to technology adoption & Conduct stakeholder engagement sessions from the start to understand concerns, resistance points, and pathways to trust and adoption. \\ 
\hline
Data quality and heterogeneity across centres & Deliver open-access generative AI models to enhance, calibrate, and harmonize image data and AI predictions across settings. \\ 
\hline
\end{tabularx}
\caption{Mitigation Plan Detailing Strategies to Address Identified Risks and Challenges}

\label{tab:mitigation}
\end{table}
Among all the previously discussed challenges, one of the most critical issues encountered during the implementation of Flower was the lack of stability in the connections established through the framework. Connection failures were frequent and disruptive, making it difficult to maintain a reliable and continuous training process. A fundamental limitation of Flower is that it requires the communication channel to remain open at all times; if the connection is interrupted, it is not always possible to reconnect and resume training. This inherent instability complicates the FL workflow, especially in low-connectivity environments, where network fluctuations are more prevalent.

Furthermore, although Flower is designed to use HTTPS with SSL certificates, it lacks a native implementation for preserving the identity and authentication of connected users. This absence of robust authentication mechanisms poses a significant security risk, as it increases the vulnerability of the communication channel to unauthorized access or malicious intrusions. The lack of authentication also increases the risk of data interception, man-in-the-middle attacks, and unauthorized participation in the FL process, potentially compromising the integrity of the training pipeline.

Integrating strong authentication mechanisms—such as multi-factor authentication (MFA), token-based verification, or public-key cryptography—would significantly enhance the security of data transmission and prevent unauthorized access. Additionally, implementing an end-to-end encryption protocol would further safeguard model updates and prevent potential data leaks, ensuring that even if an attacker gains access to the communication channel, the transmitted information remains inaccessible and unreadable.

Given these concerns, improving the reliability of connections and strengthening security measures should be prioritized in future iterations of Flower-based federated learning systems, particularly when applied to sensitive fields like healthcare, where privacy and data integrity are paramount.

Another fundamental aspect that must be addressed is the complexity of using the code as a client, particularly in settings where specialized technical personnel are not available. In many hospitals, for example, IT staff are primarily trained to manage the hospital’s daily operational and administrative systems, and they may not have the necessary expertise to handle AI model deployment, Docker containerization, or other advanced machine-learning infrastructure tasks. This lack of technical proficiency poses a significant barrier to the widespread adoption of federated learning, as setting up and maintaining the system often requires manual intervention, complex configuration, and troubleshooting, which may be beyond the capabilities of standard IT teams in clinical environments. One possible approach is to offer training courses to improve technical proficiency among IT personnel. However, this solution is not highly scalable and requires a significant investment of time, effort, and resources. Moreover, even after undergoing training, hospital IT staff would still need to dedicate additional time to keeping up with software updates, security patches, and new AI deployment methodologies, which may divert attention from their core responsibilities in hospital operations.

Another challenge is that staff turnover is common in many institutions, meaning that even if a team is trained, knowledge retention over time may be limited. New hires would require continuous onboarding and upskilling, making training alone an unsustainable long-term strategy for federated learning adoption.

For these reasons, a more practical and scalable solution is to develop intuitive, automated tools that minimize the technical burden on local IT teams. A well-designed graphical user interface (GUI) could significantly reduce the complexity of deploying federated learning models, allowing hospital staff to upload datasets, authenticate users, and monitor training processes without requiring advanced technical expertise. Additionally, automated deployment scripts and containerized environments (e.g., pre-configured Docker images) could further simplify the setup process, enabling seamless integration with existing hospital infrastructure.

By prioritizing usability and automation, federated learning can become a more accessible and efficient tool for healthcare institutions, ensuring that AI-powered solutions can be adopted without placing excessive demands on non-specialized personnel.

%To bridge this gap, it is crucial to develop a user-friendly interface that not only simplifies the deployment process but also enhances security and usability. Such an interface should include an intuitive authentication system, ensuring that only authorized personnel can access the platform. Additionally, it should provide a streamlined workflow for hospitals and research centers to easily upload their datasets, integrate them into the training pipeline, and monitor the learning process without requiring extensive technical knowledge.

Incorporating low-code or no-code solutions would allow  healthcare institutions with limited AI expertise to participate in federated learning networks without relying on external specialists. Implementing these usability improvements would significantly lower the entry barrier, making AI more accessible, scalable, and practical for real-world applications, particularly in low-resource settings even where AI deployment expertise is scarce.

\begin{figure}
    \centering
    \includegraphics[width=0.8\linewidth]{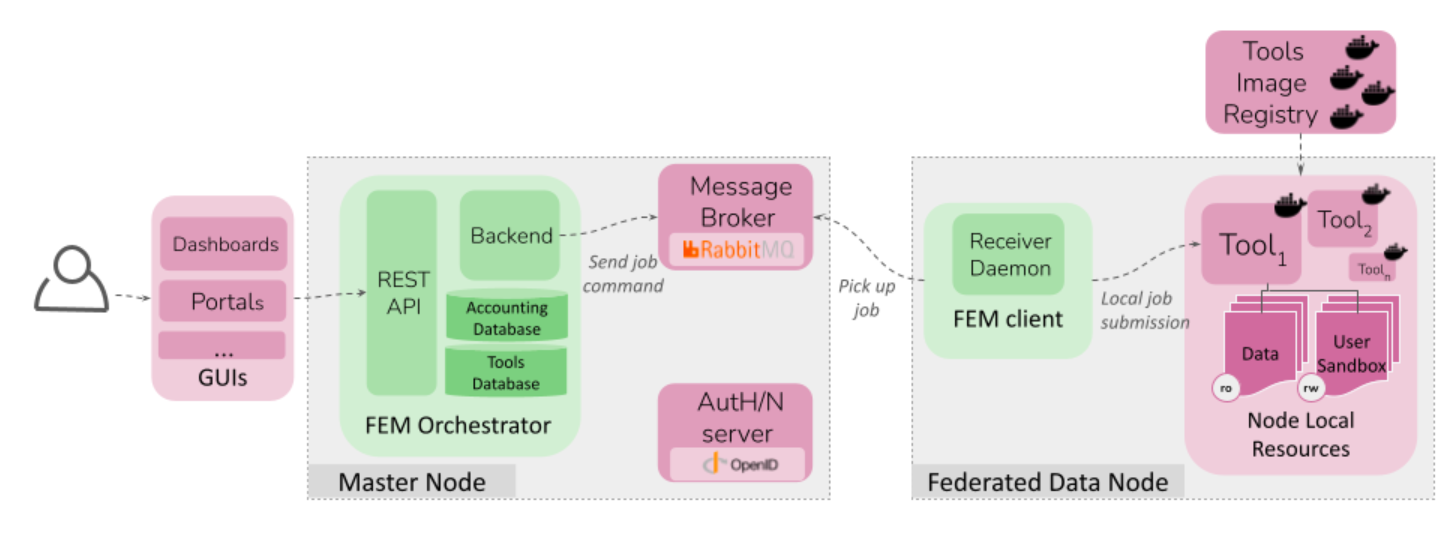}
    \caption{Architecture proposed}
    \label{fig:archit}
\end{figure}

\section{Bibliography}
%Bibliography
\bibliographystyle{unsrt}  
\bibliography{references}  
[1] H. B. McMahan, Eider Moore, Daniel Ramage, Seth Hampson, and Blaise Agüera y Arcas. Communication-efficient learning of deep networks from decentralized data. In International Conference on Artificial Intelligence and Statistics, 2016. 

[2] Tian Li, Anit Kumar Sahu, Ameet Talwalkar, and Virginia Smith. Federated learning: Challenges, methods, and future directions. IEEE Signal Processing Magazine, 37(3):50–60, 2020. 

[3] M. F. Sohan and A. Basalamah. A systematic review on federated learning in medical image analysis. IEEE Access, 11:28628–28644, 2023. 

[4] M. Moshawrab, M. Adda, A. Bouzouane, H. Ibrahim, and A. Raad. Reviewing federated learning aggregation algorithms; strategies, contributions, limitations and future perspectives. Electronics, 12(10):2287, 2023.

[5] I. Dayan, H. R. Roth, A. Zhong, A. Harouni, A. Gentili, A. Z. Abidin, A. Liu, A. Beardsworth Costa, B. J. Wood, C. S. Tsai, et al. Federated learning for medical imaging: Our experience and next steps. Journal of the American College of Radiology, 18(9):1213–1221, 2021.

[6] Sabina B. van Rooij, Muriel van der Spek, Arthur van Rooijen, and Henri Bouma. Privacy-preserving federated learning with various computer-vision tasks for security applications. In Henri Bouma, Judith Dijk, Radhakrishna
Prabhu, Robert James Stokes, and Yitzhak Yitzhaky, editors, Artificial Intelligence for Security and Defence Applications, volume 12742, page 1274205. International Society for Optics and Photonics, SPIE, 2023. 9 

[7] Q. Yang, Y. Liu, T. Chen, and Y. Tong. Federated machine learning: Concept and applications. ACM Transactions on Intelligent Systems and Technology (TIST), 10(2):1–19, 2019. 

[8] A. Linardos, K. Kushibar, S. Walsh, P. Gkontra, and K. Lekadir. Federated learning for multi-center imaging diagnostics: A simulation study in cardiovascular disease. Scientific Reports, 12(1):3551, 2022. 
 
[9] M. H. Rehman, W. Hugo Lopez Pinaya, P. Nachev, J. T. Teo, S. Ourselin, and M. J. Cardoso. Federated learning for medical imaging radiology. British Journal of Radiology, 96(1150):20220890, 2023. 

[10] A. A. S. Soltan, A. Thakur, J. Yang, A. Chauhan, L. G. D’Cruz, P. Dickson, M. A. Soltan, D. R. Thickett, D. W. Eyre, T. Zhu, and D. A. Clifton. A scalable federated learning solution for secondary care using low-cost microcomputing: Privacy-preserving development and evaluation of a covid-19 screening test in uk hospitals. The Lancet Digital Health, 6(2):e93–e104, 2024. 

[11] Jie Wen, Zhixia Zhang, Yang Lan, Zhihua Cui, Jianghui Cai, and Wensheng Zhang. A survey on federated learning: challenges and applications. International Journal of Machine Learning and Cybernetics, 14(2):513–535, 2023. 

[12] Chen Zhang, Yu Xie, Hang Bai, Bin Yu, Weihong Li, and Yuan Gao. A survey on federated learning. Knowledge- Based Systems, 216:106775, 2021.

[13] Li Li, Yuxi Fan, Mike Tse, and Kuo-Yi Lin. A review of applications in federated learning. Computers Industrial Engineering, 149:106854, 2020.

[14] Sannara EK, François PORTET, Philippe LALANDA, and German VEGA. A federated learning aggregation algorithm for pervasive computing: Evaluation and comparison. In 2021 IEEE International Conference on Pervasive Computing and Communications (PerCom), pages 1–10, 2021. 

[15] Abejide Ade-Ibijola and Chinedu Okonkwo. Artificial Intelligence in Africa: Emerging Challenges, pages 101–117. Springer International Publishing, Cham, 2023. 

[16] Peter Kairouz, H. Brendan McMahan, Brendan Avent, Aurélien Bellet, Mehdi Bennis, Arjun Nitin Bhagoji, Kallista Bonawitz, Zachary Charles, Graham Cormode, Rachel Cummings, Rafael G. L. D’Oliveira, Hubert Eichner, Salim El Rouayheb, David Evans, Josh Gardner, Zachary Garrett, Adrià Gascón, Badih Ghazi, Phillip B.
Gibbons, Marco Gruteser, Zaid Harchaoui, Chaoyang He, Lie He, Zhouyuan Huo, Ben Hutchinson, Justin Hsu, Martin Jaggi, Tara Javidi, Gauri Joshi, Mikhail Khodak, Jakub Konecný, Aleksandra Korolova, Farinaz
Koushanfar, Sanmi Koyejo, Tancrède Lepoint, Yang Liu, Prateek Mittal, Mehryar Mohri, Richard Nock, Ayfer Özgür, Rasmus Pagh, Hang Qi, Daniel Ramage, Ramesh Raskar, Mariana Raykova, Dawn Song, Weikang Song, Sebastian U. Stich, Ziteng Sun, Ananda Theertha Suresh, Florian Tramèr, Praneeth Vepakomma, Jianyu Wang, Li Xiong, Zheng Xu, Qiang Yang, Felix X. Yu, Han Yu, and Sen Zhao. Advances and open problems in federated
learning. Foundations and Trends® in Machine Learning, 14(1–2):1–210, 2021.

[17] Pengrui Liu, Xiangrui Xu, and Wei Wang. Threats, attacks and defenses to federated learning: issues, taxonomy and perspectives. Cybersecurity, 5(1):4, 2022. 

[18] D. Yao, W. Pan, Y. Dai, Y. Wan, X. Ding, H. Jin, et al. Local–global knowledge distillation in heterogeneous federated learning with non-iid data. arXiv, 2021.

[19] Y. He, Y. Chen, X. Yang, H. Yu, Y.-H. Huang, and Y. Gu. Learning critically: Selective self-distillation in federated learning on non-iid data. IEEE Transactions on Big Data, July 2022. 

[20] Y. He, Y. Chen, X. Yang, Y. Zhang, and B. Zeng. Class-wise adaptive self distillation for heterogeneous federated learning. In Proceedings of the 36th AAAI Conference on Artificial Intelligence, volume 22, pages 1–6, 2022. 

[21] X. Dong, S. Q. Zhang, A. Li, and H. T. Kung. Spherefed: Hyperspherical federated learning. In Computer Vision—ECCV 2022, pages 165–184. Springer, Cham, Switzerland, 2022. 

[22] AAS Soltan, A Thakur, J Yang, A Chauhan, LG D’Cruz, P Dickson, MA Soltan, DR Thickett, DW Eyre, T Zhu, and DA Clifton. A scalable federated learning solution for secondary care using low-cost microcomputing: privacy-preserving development and evaluation of a covid-19 screening test in uk hospitals. The Lancet Digital Health, 6(2):e93–e104, Feb 2024. 

[23] Bingjie Yan, Jun Wang, Jieren Cheng, Yize Zhou, Yixian Zhang, Yifan Yang, Li Liu, Haojiang Zhao, Chunjuan Wang, and Boyi Liu. Experiments of federated learning for covid-19 chest x-ray images. In Xingming Sun, Xiaorui Zhang, Zhihua Xia, and Elisa Bertino, editors, Advances in Artificial Intelligence and Security, pages
41–53, Cham, 2021. Springer International Publishing.

[24] Tawsifur Rahman, Amith Khandakar, Muhammad A. Kadir, Khandaker R. Islam, Khandaker F. Islam, Zaid B. Mahbub, Mohamed Arselene Ayari, and Muhammad E. H. Chowdhury. Reliable tuberculosis detection using chest x-ray with deep learning, segmentation and visualization. IEEE Access, 8:191586–191601, 2020. 10

[25] D. J. Beutel, T. Topal, A. Mathur, X. Qiu, J. Fernandez-Marques, Y. Gao, L. Sani, H. L. Kwing, T. Parcollet,
P. P. B. Gusmão, and N. D. Lane. Flower: A friendly federated learning research framework. arXiv preprint, 2020.

[26] J. Ansel, E. Yang, H. He, N. Gimelshein, A. Jain, M. Voznesensky, B. Bao, P. Bell, D. Berard, E. Burovski, et al. Pytorch 2: Faster machine learning through dynamic python bytecode transformation and graph compilation. In Proceedings of the 29th ACM International Conference on Architectural Support for Programming Languages and Operating Systems (ASPLOS ’24), volume 2. ACM, April 2024. 

[27] W. Falcon and The PyTorch Lightning Team. Pytorch lightning (version 1.4) [computer software], March 2019. 

[28] K. He, X. Zhang, S. Ren, and J. Sun. Deep residual learning for image recognition. In Proceedings of the IEEE Conference on Computer Vision and Pattern Recognition, pages 770–778, 2016. \\

\end{document}